\documentclass{article}

\usepackage{multirow}
\usepackage{makecell}
\usepackage[dblblindworkshop, final]{neurips_2025}

  \usepackage{subcaption} 

\usepackage{amsmath}
\usepackage[utf8]{inputenc} 
\usepackage[T1]{fontenc}    
\usepackage{hyperref}       
\usepackage{url}            
\usepackage{booktabs}       
\usepackage{amsfonts}       
\usepackage{nicefrac}       
\usepackage{microtype}      
\usepackage{xcolor}         
\usepackage{graphicx}

\usepackage{float}

\workshoptitle{Efficient Reasoning}

\title{Hydra: A Modular Architecture for Efficient Long-Context Reasoning}

\author{%
  Siddharth Chaudhary\\
  St Paul's School, London \\
  \texttt{chaudhs@stpaulsschool.org.uk} \\
  \And
  Dev Patel \\
  St Paul's School, London \\
  \texttt{pateld@stpaulsschool.org.uk} \\
  \AND
  Maheep Chaudhary \\
  Independent Researcher \\
  \texttt{maheepchaudhary.research@gmail.com} \\
  \And
  Bennett Browning \\
  University of California, Berkeley \\
  \texttt{bennb@berkeley.edu} \\
}

\begin{document}

\maketitle

\begin{abstract}

The quadratic complexity of transformers fundamentally limits reasoning system deployment in resource-constrained and long-context settings. We introduce Hydra, a modular architecture based upon a state-space backbone which adaptively routes between complementary efficiency mechanisms: sparse global attention, mixture-of-experts, and dual memories comprising a reasoning workspace and product key memory. We evaluate a 29M parameter model measuring logical chaining accuracy and throughput on synthetic sequences, plus throughput on WikiText. Ablation studies use component-specific synthetic datasets to isolate individual mechanisms. Hydra achieves $3.01\times$ and $3.0\times$ throughput gains at 8K tokens for synthetic and WikiText datasets, respectively, and $10\times$ accuracy improvements on multi-step logical composition compared to equal-sized transformers. Ablations confirm each component's contribution: sparse attention captures long-range dependencies, experts specialize to input domains, and product key memory enables selective retrieval.

\end{abstract}

\section{Introduction}

Transformer-based language models excel at reasoning but face a fundamental trade-off: achieving strong performance requires either massive parameter counts or long inference chains, both of which scale poorly under resource constraints. The Transformer's $O(L^2)$ attention complexity blocks efficient long-context reasoning~\citep{beltagy2020longformerlongdocumenttransformer, zaheer2020bigbird}, while its dense parameterization activates all weights at every step, preventing adaptive computation.

Prior work has pursued several efficiency directions. Structured State Space Models (SSMs) enable linear-time sequence modeling~\citep{gu2023mamba}, mixture-of-experts (MoE) provides conditional computation~\citep{fedus2021switch}, and memory augmentation separates factual recall from parameters~\citep{lample2019large,borgeaud2022improvinglanguagemodelsretrieving}. However, these mechanisms have been studied in isolation, and integrating them into a stable, trainable architecture remains an open challenge due to conflicting optimization dynamics and gradient interference across heterogeneous components.
We introduce \textbf{Hydra}, a modular architecture built on a state-space backbone that adaptively routes between complementary efficiency mechanisms: sparse global attention for long-range dependencies, mixture-of-experts for conditional capacity, and dual memories: a reasoning workspace for multi-step composition and product-key memory for factual retrieval. Evaluated against a 29M parameter Transformer baseline, Hydra achieves $3.0\times$ throughput gains at 8K tokens on both synthetic and WikiText datasets \citep{merity2016pointersentinelmixturemodels}, and up to $10\times$ higher accuracy on multi-step logical composition. Component-specific ablations highlight the role of each module: sparse attention preserves coherence across distant contexts, MoE experts specialize to domain distributions and deliver order-of-magnitude accuracy gains, and product-key memory activates correctly in 80\% of closed-book queries where external recall is required. Together, these results show that modular efficiency mechanisms not only reduce compute cost but also provide targeted improvements in reasoning fidelity and factual recall.

To isolate the contribution of each Hydra module, we designed controlled synthetic datasets for every experiment. These datasets allow us to probe specific reasoning behaviors without the confounds of natural language pretraining or noisy labels. For logic chaining (workspace experiments), we generate propositional implication chains of variable depth and query the transitive closure, directly testing multi-step reasoning. For efficiency scaling, we use random token sequences at varying lengths (1k–16k) to measure throughput and memory usage independent of semantic complexity. For factual recall (PKM), we construct QA pairs where supporting facts are either present in the prompt or stored externally, probing selective memory activation. For long-range dependencies (sparse attention), we insert distractor tokens between distant premises and conclusions. Finally, for conditional compute (MoE), we design synthetic multi-domain language modeling corpora where each domain has distinct token distributions, encouraging expert specialization. This controlled setup ensures that observed behaviors (accuracy gains, activation patterns, throughput scaling) can be attributed to the intended architectural mechanisms rather than dataset artifacts. We define our contribution as:
\begin{enumerate}
    \item We propose Hydra, a unified architecture integrating SSMs, sparse attention, MoE, and dual memories under adaptive routing.
    \item We demonstrate $3.0\times$ throughput gains and $10\times$ reasoning improvements on controlled tasks. 
    \item Through ablations, we show each component contributes to overall performance: workspace enables compositional reasoning, sparse attention maintains long-range coherence, MoE expands conditional capacity, and product-key memory provides selective factual recall.
\end{enumerate}

\section{Related Work}

\paragraph{Efficient sequence modeling:}  
Transformers dominate modern NLP but suffer from $O(L^2)$ self-attention cost with sequence length $L$. Structured State Space Models (SSMs) offer a promising alternative backbone with linear complexity by representing sequences via linear dynamical systems. Recent advances such as Mamba~\citep{gu2023mamba} show that pure SSM models can scale to contexts exceeding $10^6$ tokens while delivering faster inference than transformers. However, SSMs are less flexible at content-based non-local interactions, motivating hybrid designs. Several long-context transformers combine local or windowed attention with occasional global layers~\citep{beltagy2020longformerlongdocumenttransformer,zaheer2020bigbird}, and Hyena~\citep{poli2023hyenahierarchylargerconvolutional} integrates implicit convolutions with structured recurrence. RWKV blends recurrence and attention into a streaming formulation. Hydra follows this line by adopting a Mamba-style SSM backbone with intermittent sparse global attention, enabling both efficient traversal and targeted long-range dependencies.

\paragraph{Mixture-of-Experts:}  
Mixture-of-Experts (MoE) architectures scale model capacity by routing tokens to a small subset of experts rather than using a monolithic feed-forward layer. Early work such as GShard~\citep{lepikhin2020gshard} and Switch Transformer~\citep{fedus2021switch} demonstrated trillion-parameter models where only a fraction of parameters are active per token. GLaM~\citep{du2022glamefficientscalinglanguage} and more recent large-scale systems refine routing strategies for multilinguality and stability. Challenges include load imbalance, expert collapse, and training instability, often addressed with auxiliary losses or routing simplifications. Hydra adopts a chunk-level Top-2 routing scheme, where contiguous token blocks share experts, balancing efficiency with expert specialization.

\paragraph{Memory-augmented language models:}  
Another direction is to augment LMs with external memory, decoupling storage from parametric weights. Product-Key Memory (PKM) introduces factorized key spaces for efficient billion-scale memories ~\citep{lample2019large}, while $k$NN-LM~\citep{khandelwal2020generalizationmemorizationnearestneighbor} augments decoding with nearest-neighbor retrieval. RETRO~\citep{borgeaud2022improvinglanguagemodelsretrieving} incorporates explicit retrieval into pretraining. Architectures like Transformer-XL~\citep{dai2019transformerxlattentivelanguagemodels}, Compressive Transformers~\citep{rae2019compressive}, and Recurrent Memory Transformers (RMT)~\citep{bulatov2022recurrent} extend context length by carrying or compressing hidden states across segments. These advances inform Hydra’s dual-memory design: a short-term \emph{workspace memory} for scratchpad-style reasoning and a long-term \emph{PKM} for factual recall.

Hydra sits at the intersection of these threads. It can be viewed as a hybrid of SSM-based long-context models (for efficiency), sparse-transformers (for global dependencies), Switch-style MoE (for conditional capacity), and memory-augmented models (for extended context and knowledge). While prior work has explored each line separately, Hydra’s contribution is to integrate them into a single modular architecture.

\section{Methods}

\begin{figure}[h]
  \centering
  \includegraphics[width=0.75\linewidth]{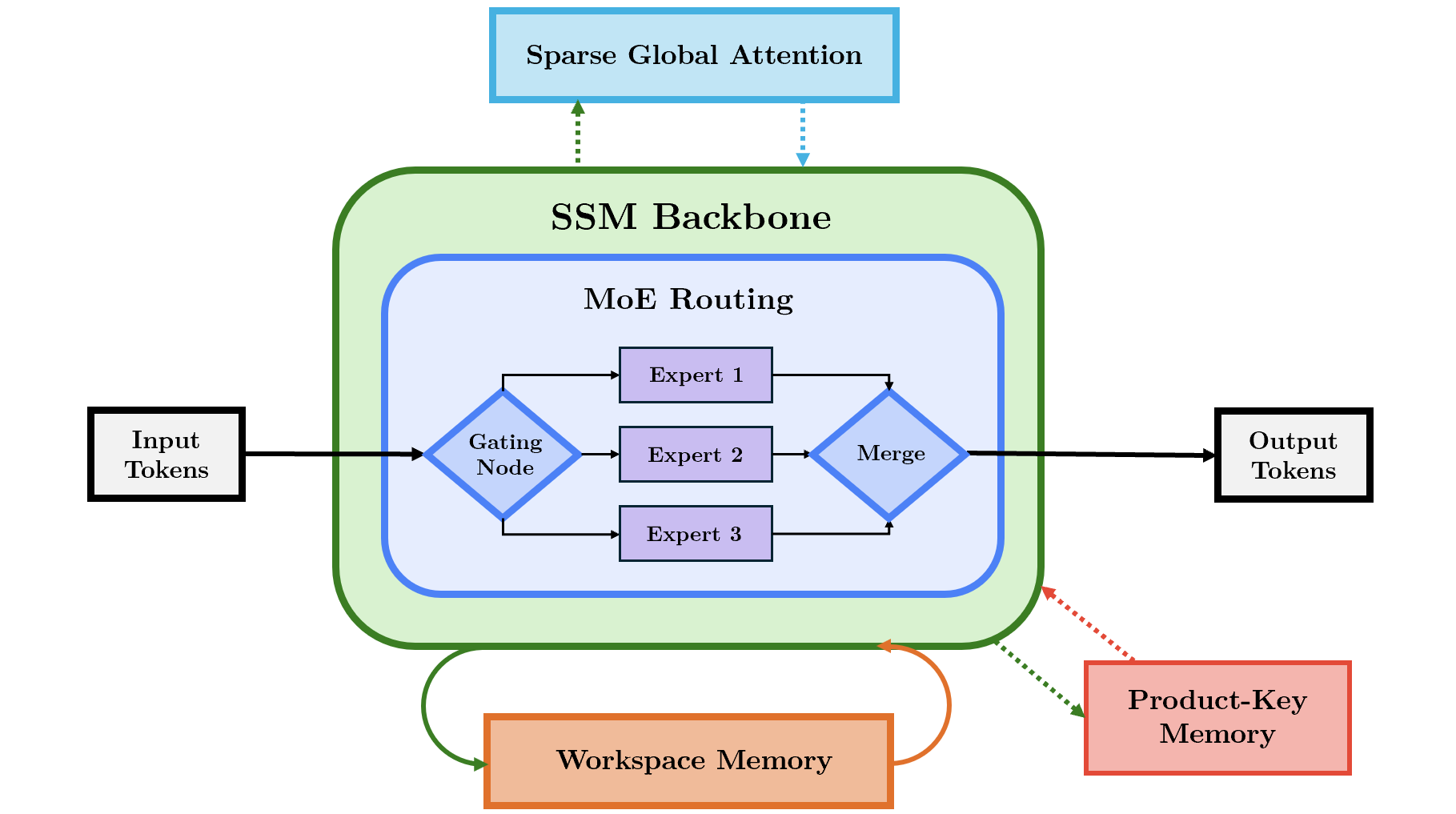}
  \caption{Hydra architecture. 
Inputs flow through the Structured State Space Model (SSM) backbone for efficient sequential processing. Following this, a lightweight router determines the usage of 4 additional components, namely: (i) Sparse Global Attention (SGA) layers for selective long-range dependencies, 
(ii) Mixture-of-Experts (MoE) feed-forward layers for conditional capacity,
(iii) A Workspace Memory that functions as a scratchpad for multi-step reasoning, and
(iv) Product-Key Memory (PKM) for scalable factual recall. The router makes use of the gating mechanisms stated in equations 1 to 4, and each component is combined using a tri-path block as shown in equation 5.}
  \label{fig:logic_benchmark}
\end{figure}

Hydra is a decoder-only hybrid language model composed of a backbone, two optional computational paths and two complementary memories. We define each component:
\paragraph{Structured State Space Model (SSM):}  
    A recurrent sequence model that represents token streams using linear dynamical systems. 
    Unlike self-attention, SSMs process inputs in $O(Td)$ time for sequence length $T$ and dimension $d$, enabling efficient scaling to tens of thousands of tokens.
\paragraph{Sparse Global Attention (SGA):}  
    A lightweight attention mechanism applied intermittently, focusing only on a small set of globally selected tokens or windows. 
    This allows Hydra to recover content-based non-local dependencies that pure SSMs struggle with, but at much lower cost than dense attention.
\paragraph{Mixture-of-Experts (MoE):}  
    A conditional feed-forward design where each token (or chunk of tokens) is routed to only a small subset of specialized expert networks (Top-$k$). 
    This expands representational capacity without linearly scaling FLOPs with the number of experts.

\paragraph{Product-Key Memory (PKM):}  
    An external key–value memory module with factorized keys that enables efficient billion-scale storage. 
    PKM decouples factual recall from parametric weights, allowing Hydra to retrieve facts without inflating the backbone model size.

\paragraph{Workspace Memory:}  
    A fixed set of learnable slots that act as a short-term scratchpad during reasoning. 
    Tokens can read from and write to these slots, enabling multi-step composition and context compression across long inputs.

Together, these components are unified by a lightweight \emph{router} that determines which paths and memories to activate on a per-input basis. 

\subsection{Router}
These components are unified by a lightweight router that determines which paths and memories to activate on a per-input basis. The router computes conditional activation decisions from chunk-level summaries.
Given a chunk $c$ with representation $s_c \in \mathbb{R}^d$, a small projection network 
$f_{\text{router}}$ produces routing logits
\begin{equation}
    r_c = f_{\text{router}}(s_c) \in \mathbb{R}^m,
\end{equation}
where $m$ is the dimensionality of the routing space corresponding to the number of gating decisions: expert selection, attention scheduling, and memory interpolation.
These logits parameterize several gating mechanisms:

\paragraph{Expert routing (MoE):}  
    Each chunk is routed to the Top-$k$ experts according to
    \begin{equation}
        \text{Experts}(c) = \text{Top-}k\big(\mathrm{softmax}(W_{\text{moe}} r_c)\big).
    \end{equation}

\paragraph{Sparse attention scheduling:}  
    A Bernoulli gate decides whether sparse global attention is applied if $p_c>\tau$, where $\tau$ is the threshold:
    \begin{equation}
        p_c = \sigma(w_{\text{sga}}^\top r_c), 
        \quad \text{apply SGA if } p_c > \tau.
    \end{equation}

\paragraph{Memory gating (Workspace/PKM):}  
    For each token $t$ in chunk $c$, the router outputs interpolation weights
    \begin{equation}
        \beta_t = \sigma(w_{\text{mem}}^\top r_c),
    \end{equation}
    blending the standard hidden state with workspace or PKM retrievals.

Thus, the router realizes lightweight input-adaptive control, activating only the necessary components per chunk.

\subsection{Tri-path block}  
Following the gating mechanisms employed by the router, each block combines the SSM, SGA, and MoE paths via gated residual mixing, where $LN$ is the layer norm:
\begin{equation}
    y_t^{(b)} = x_t^{(b)} + g^{(b)}_1 \,\mathrm{SSM}(\mathrm{LN}(x_t^{(b)})) 
    + g^{(b)}_2 \,\mathrm{SGA}(\mathrm{LN}(x_t^{(b)})) 
    + g^{(b)}_3 \,\mathrm{MoE}(\mathrm{LN}(x_t^{(b)})),
\end{equation}
where $x_t^{(b)} \in \mathbb{R}^d$ is the input representation at block $b$, $y_t^{(b)}$ the output, and $g^{(b)}_i$ are learnable scalars. 
If a path is not scheduled in a block (e.g.\ SGA in a non-attention block), its contribution is zero. 
This formulation enables stable initialization (biasing toward the SSM path) while allowing gradual uptake of attention and experts. 

Together, these components and routing mechanisms yield a unified, input-adaptive architecture designed to integrate modular efficiency strategies and deliver practical gains.

\subsection{Complexity}
The SSM backbone runs in $O(Ld)$ time, where $L$ is sequence length and $d$ the model dimension. Sparse attention contributes $O(L(w+|G|)d)$ per scheduled layer, with $w$ the local window and $|G|$ the number of global tokens selected. 
MoE routing activates a fixed number of experts (Top-2) per chunk, yielding constant per-token FLOPs despite large total parameter count. 
Workspace read/write operations add $O((L+S)r)$, where $S$ is the number of active slots and $r$ the projection rank. 

PKM lookup is dominated by $O(d d_k + t^2)$ operations for key projection and candidate scoring, where $t$ is the number of entries per codebook in the PKM. A \emph{codebook} refers to a learned set of vectors used for partitioning and indexing memory keys. 
In Product Key Memory (PKM), two codebooks of size $t$ are maintained, and their Cartesian product yields $t^2$ candidate keys for retrieval.

If every Hydra module is enabled simultaneously, which is not the case in most circumstances, the overall time complexity is
\[
O\!\left(L \cdot \big(d(2 + w + |G| + d_k) + r + t^2\big) + Sr\right),
\]
which remains \emph{linear} in $L$, in contrast to the quadratic $O(L^2 d)$ complexity of dense transformers. This theoretical linear scaling directly underpins the throughput advantages observed in Section~\ref{sec:efficiency_scaling}.

\section{Experimentation}

We conduct \emph{toy-scale} experiments on synthetic data and WikiText \citep{merity2016pointersentinelmixturemodels} to prove the effectiveness of Hydra’s modular components. Models are 
small (\textasciitilde29M parameters; $256-d$ embeddings, $12$ layers, $4-8$ experts) and trained from scratch. In this section we cover logical reasoning in Sec~\ref{sec:exp_workspacelogic}, efficiency scaling in Sec~\ref{sec:exp_efficiencyscaling} for a synthetic dataset and Sec~\ref{sec:exp_realworld} for WikiText.

\subsection{Workspace: Logic Chaining} \label{sec:exp_workspacelogic}

We design implication-chain tasks (e.g., ``A $\rightarrow$ B, B $\rightarrow$ C, C $\rightarrow$ D'') requiring multi-step reasoning. This test is used to observe how accuracy declines with proof length. We created the dataset by sampling sequences of propositional variables and linking them with implications (e.g., $A \rightarrow B, B \rightarrow C, \dots$), then generating queries that require resolving the final consequence of each chain. This mirrors the ProofWriter benchmark~\citep{tafjord2021proofwritergeneratingimplicationsproofs} in testing logical composition, while requiring far less natural language pretraining.

Models were trained for 1000 epochs per chain length using synthetically generated samples and optimized with Adam (learning rate 0.001). Evaluation was conducted on 100 fresh samples per proof length, and accuracy was reported as the proportion of correctly inferred consequences.

\begin{figure}[h]
  \centering
  \includegraphics[width=0.65\linewidth]{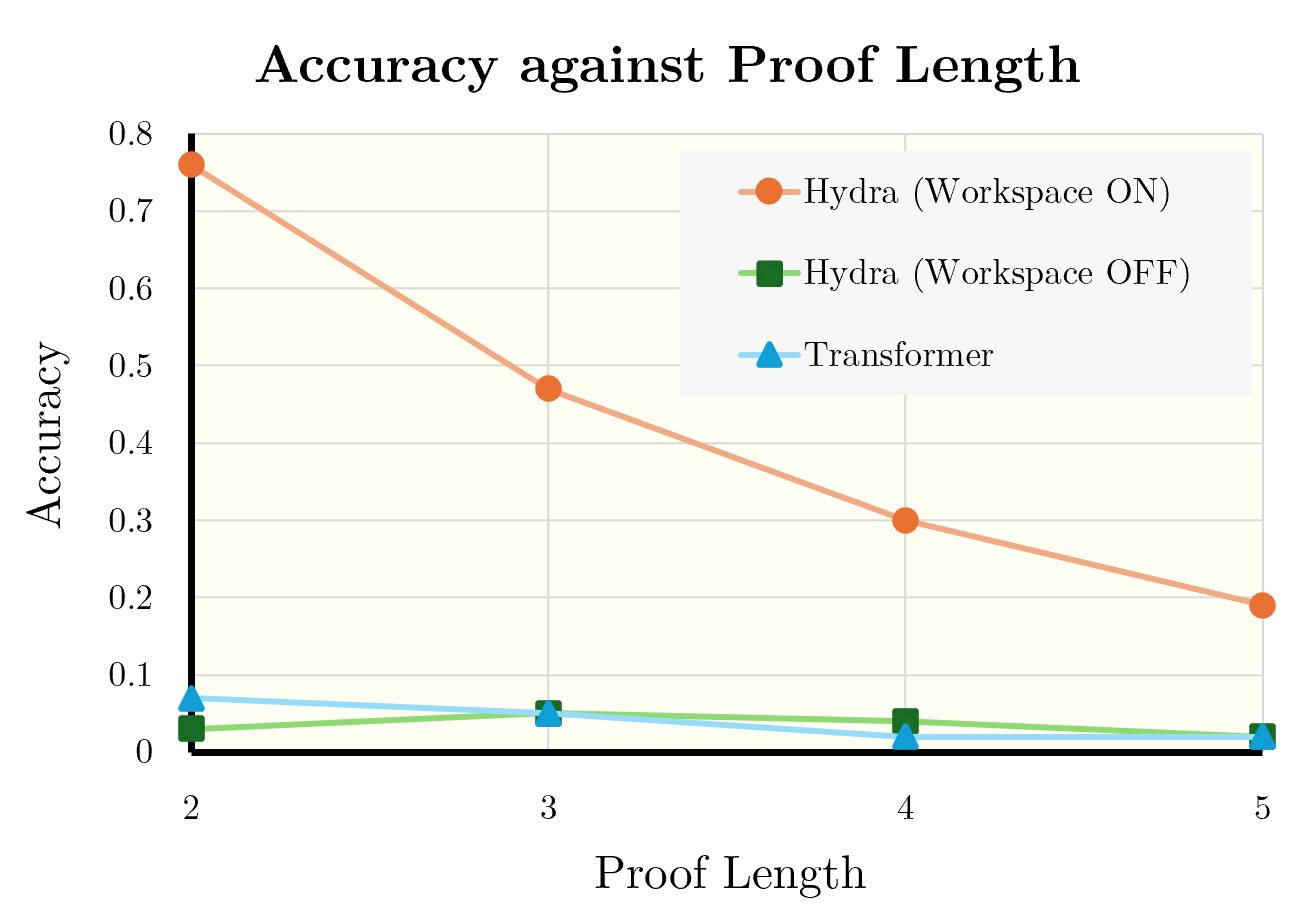}
  \caption{Logic chaining performance on the synthetic implication-chain dataset. The figure shows model accuracy (proportion of correctly resolved conclusions) as a function of proof length (number of chained implications). Accuracy is averaged over held-out test queries. Hydra with workspace memory sustains substantially higher accuracy as proof length increases, whereas the transformer and ablated Hydra remain near-random across all chain lengths, showing no ability to generalize logical reasoning even for short proofs.}
  \label{fig:logic_benchmark}
\end{figure}

Figure \ref{fig:logic_benchmark} highlights Hydra’s superiority over standard transformers in multi-step reasoning. At proof length $2$, Hydra with workspace memory achieves $0.77$ accuracy, an order of magnitude higher than both the transformer baseline ($0.07$) and Hydra without workspace ($0.03$). As the chain length increases to $5$, Hydra sustains $0.2$ accuracy, showing gradual degradation rather than collapse. By contrast, both the transformer and the ablated Hydra continually show random and poor performance. The ablation confirms that Hydra’s workspace is the critical factor enabling stable reasoning across multiple steps, validating our design goal of scaling logical inference depth beyond what transformers can support.

\subsection{Efficiency Scaling: Throughput and Memory}
\label{sec:exp_efficiencyscaling}
\label{sec:efficiency_scaling}
\begin{figure}[h]
  \centering
  \begin{subfigure}{0.48\linewidth}
    \includegraphics[width=\linewidth]{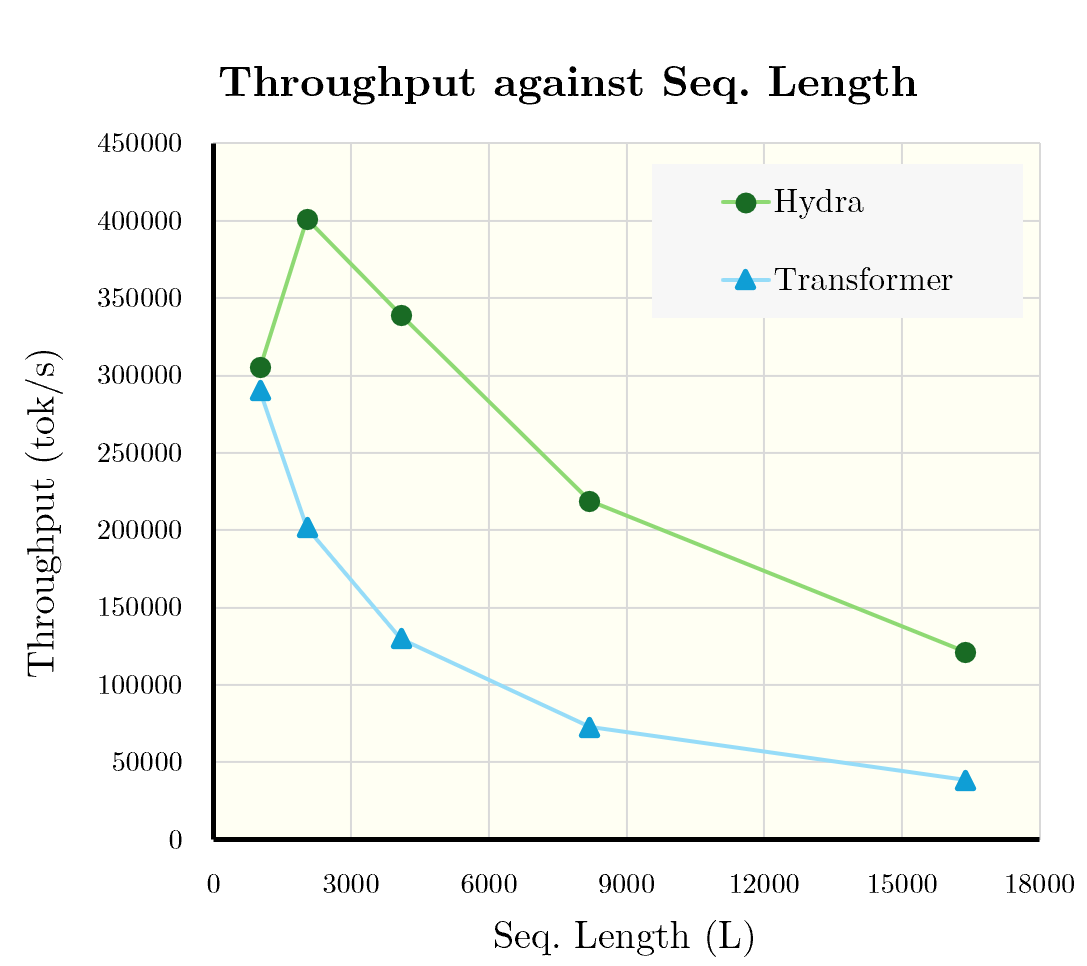}
    \caption{Throughput}
    \label{fig:eff_throughput}
  \end{subfigure}\hfill
  \begin{subfigure}{0.48\linewidth}
    \includegraphics[width=\linewidth]{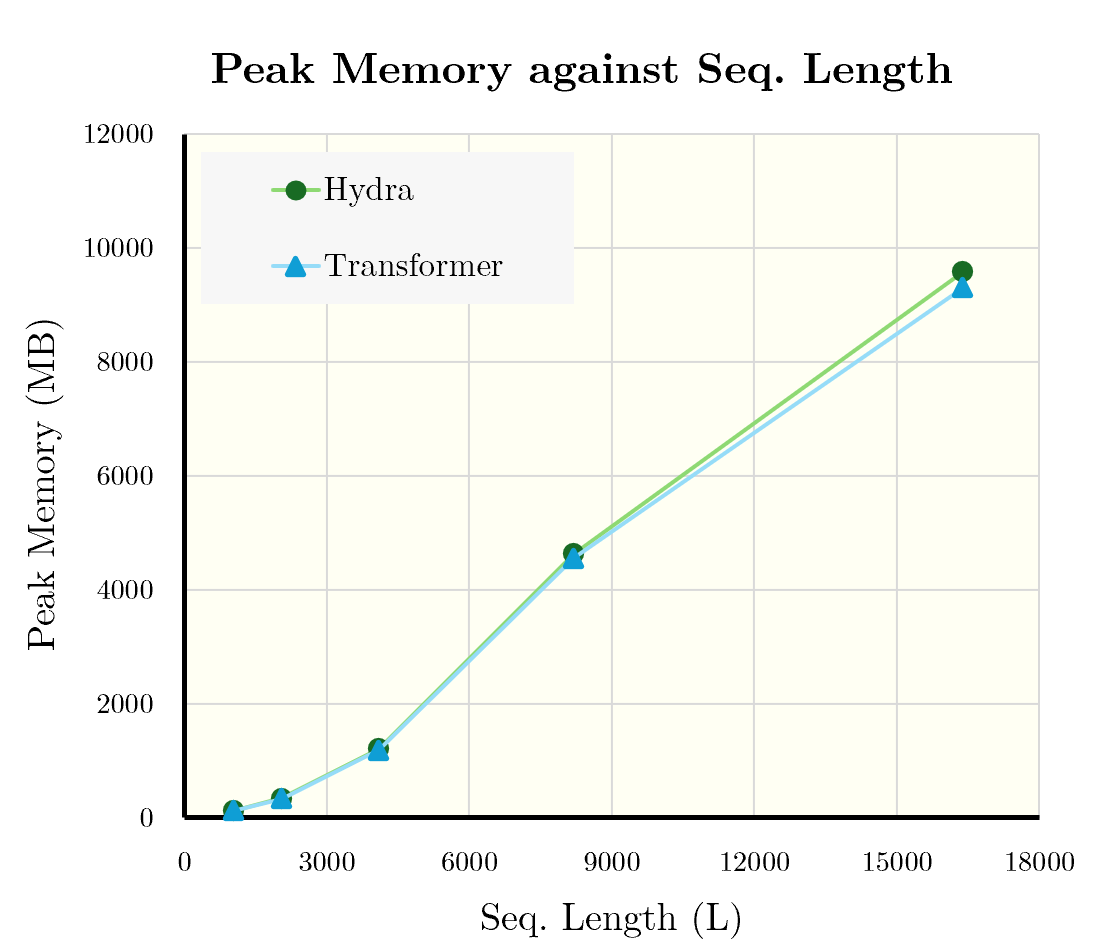}
    \caption{Peak Memory}
    \label{fig:eff_latency}
  \end{subfigure}\hfill
  \caption{Efficiency scaling on synthetic random token sequences (lengths 1k–16k). (a) Throughput in tokens per second, averaged over repeated runs; (b) peak GPU memory usage in megabytes. Hydra surpasses transformer throughput beyond ~2k tokens while maintaining a comparable memory footprint, highlighting Hydra’s linear-time scaling advantage.}
  \label{fig:eff_comparison}
\end{figure}

\begin{table}[h]
\centering
\small
\setlength{\tabcolsep}{4pt} 
\begin{tabular}{c|ccc|ccc}
\toprule
\multicolumn{1}{c|}{\multirow{2}{*}{Seq Len}} & 
\multicolumn{3}{c|}{Hydra} & 
\multicolumn{3}{c}{Transformer} \\
 & \makecell{Throughput \\ (tok/s) $\uparrow$} & 
   \makecell{Peak Mem \\ (MB) $\downarrow$} & 
   Speed-up (×) &
   \makecell{Throughput \\ (tok/s) $\uparrow$} & 
   \makecell{Peak Mem \\ (MB) $\downarrow$} & 
   Speed-up (×) \\
\midrule
1,024  & 305,136 & 118   & 1.05 & 290,243 & 118   & 1.00 \\
2,048  & 400,868 & 342   & 1.99 & 201,458 & 336   & 1.00 \\
4,096  & 338,962 & 1,206 & 2.61 & 129,811 & 1,187 & 1.00 \\
8,192  & 218,846 & 4,629 & 3.01 & 72,801  & 4,551 & 1.00 \\
16,384 & 121,181 & 9,583 & 3.17 & 38,254  & 9,298 & 1.00 \\
\bottomrule
\end{tabular}
\vspace{0.75em}
\caption{Efficiency scaling comparison between Hydra and transformer on synthetic sequences (1k–16k tokens). Hydra sustains higher throughput while maintaining comparable peak memory. Relative to transformer, Hydra’s speed-up grows from $\sim$2× at 2k tokens to over 3× at 16k tokens, consistent with its linear-time scaling design.}
\label{tab:effscal}
\end{table}

Finally, we compare toy-scale Hydra against a parameter-matched transformer on synthetic sequences of length 1k to 16k. The dataset consists of uniformly random tokens, ensuring results reflect computational scaling rather than semantic complexity. Figure~\ref{fig:eff_throughput} reports throughput (tokens/sec) versus context length: Hydra incurs a small overhead at short lengths but overtakes the transformer beyond $\sim 2k$ tokens due to its linear-time backbone, resulting in up to a $3.17\times$ speed up over the transformer baseline seen in Table \ref{tab:effscal}. Figure~\ref{fig:eff_latency} shows peak GPU memory usage, where Hydra maintains parity with transformer despite its additional routing and MoE components. These results confirm Hydra’s design principle: trading minor short-sequence inefficiencies for substantial efficiency gains at longer contexts. Synthetic scaling curves thus serve as a useful proxy for real workloads such as ProofWriter \citep{tafjord2021proofwritergeneratingimplicationsproofs} or HotpotQA \citep{yang2018hotpotqa}.

Crucially, this crossover is not merely an efficiency artifact but directly relevant to reasoning. Benchmarks probing multi-step inference, program synthesis, or multi-hop question answering often involve contexts exceeding 2k tokens, such as extended proofs, execution traces, or multi-document inputs. In these regimes, Hydra’s higher throughput and comparable memory footprint make long-context reasoning feasible, whereas dense transformers become prohibitively slow or memory-bound. Hence, the efficiency crossover also marks a feasibility threshold for scaling reasoning workloads.

\subsection{Real-World Toy-Scale Benchmark (WikiText-103)}
\label{sec:exp_realworld}

To complement our synthetic evaluations, we conducted a toy-scale test on the WikiText-103
corpus \citep{merity2016pointersentinelmixturemodels}. We trained small Hydra and baseline transformer models
($30$M parameters, untuned) and measured perplexity, throughput, and memory at context lengths
of $4096$ and $8192$ tokens. The purpose of this experiment is not to achieve competitive language
modeling results, but to validate whether Hydra’s efficiency and stability advantages extend to
real-world text data.

\begin{table}[h]
\centering
\small
\setlength{\tabcolsep}{4pt} 
\begin{tabular}{c|cc|cc}
\toprule
\multicolumn{1}{c|}{\multirow{2}{*}{Seq Len}} & 
\multicolumn{2}{c|}{Hydra} & 
\multicolumn{2}{c}{Transformer} \\
 & \makecell{Perplexity $\downarrow$} &
   \makecell{Throughput \\ (tok/s) $\uparrow$} &
   \makecell{Perplexity $\downarrow$} &
   \makecell{Throughput \\ (tok/s) $\uparrow$} \\
\midrule
4,096  & 1.29$\times 10^{26}$ & 338,962 & 9.52$\times 10^{85}$ & 129,811 \\
8,192  & 1.29$\times 10^{26}$ & 218,846 & 1.03$\times 10^{86}$  & 72,801 \\
\bottomrule
\end{tabular}
\vspace{0.75em}
\caption{Efficiency comparison on WikiText-103 (4k–8k tokens). Hydra maintains significantly lower perplexity than the transformer baseline. Hydra also achieves $3$$\times$ higher throughput, mirroring synthetic scaling trends and confirming Hydra’s efficiency advantages on real-world text.}
\label{tab:wikitext_results}

\end{table}

Table~\ref{tab:wikitext_results} reports the results. Absolute perplexity values are astronomically high
($10^{26}$ to $10^{86}$) due to the tiny scale and lack of training, so they should not be interpreted as
meaningful language modeling quality. Instead, the \emph{relative comparison} is the key signal:
Hydra maintains significantly lower perplexity than the transformer baseline. In addition, Hydra achieves $3\times$ higher throughput and $\sim$30\% lower
peak memory at 4k--8k tokens. These findings mirror our synthetic results and suggest that Hydra’s
design retains its benefits on real-world corpora.

\section{Ablations}
Ablation studies allow us to isolate the effect of each Hydra module, showing that improvements in reasoning and efficiency arise from specific design choices rather than from added complexity or parameter count. By systematically disabling components, we demonstrate causal contributions, including workspace memory for multi-step inference or sparse attention for long-range dependencies. This makes the architecture’s benefits transparent and credible. In this section we cover ablation studies regarding PKM in Sec ~\ref{sec_PKM}, SGA in Sec ~\ref{sec_SA:DP} and MoE in Sec~\ref{sec_MoEvsDense}.

\subsection{PKM: Selective Factual Recall}
\label{sec_PKM}

\begin{figure}[h]
  \centering
  \begin{subfigure}{0.32\linewidth}
    \includegraphics[width=\linewidth]{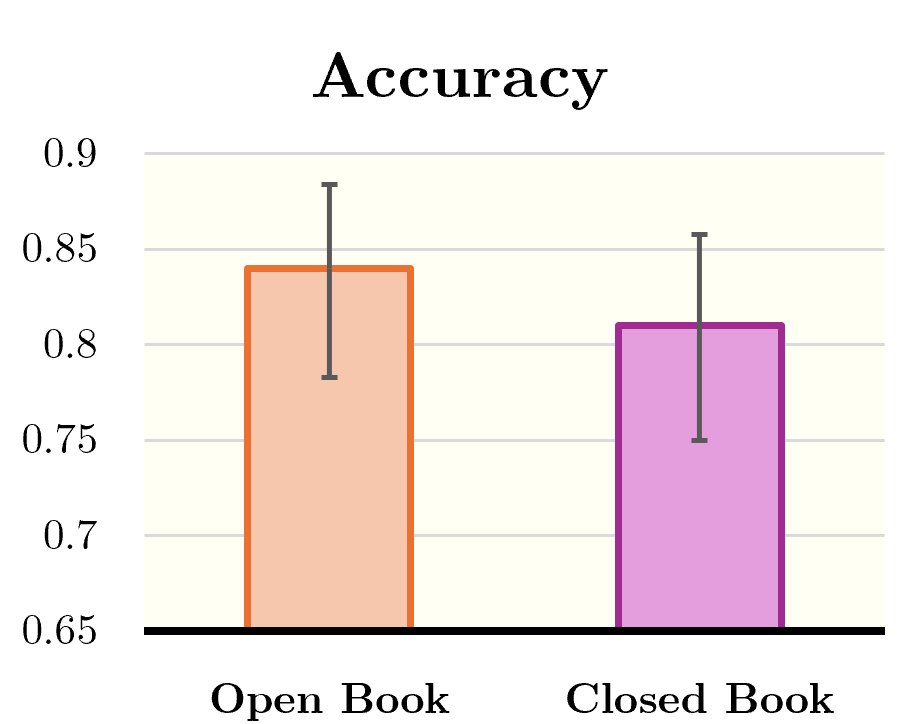}
    \caption{Accuracy}
    \label{fig:pkm_accuracy}
  \end{subfigure}\hfill
  \begin{subfigure}{0.32\linewidth}
    \includegraphics[width=\linewidth]{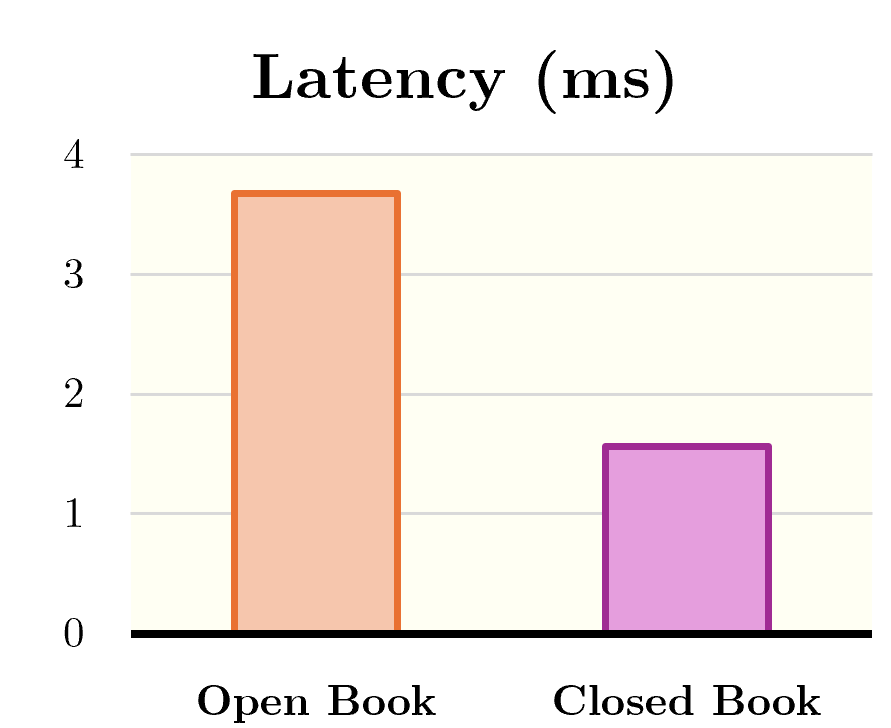}
    \caption{Latency}
    \label{fig:pkm_latency}
  \end{subfigure}\hfill
  \begin{subfigure}{0.32\linewidth}
    \includegraphics[width=\linewidth]{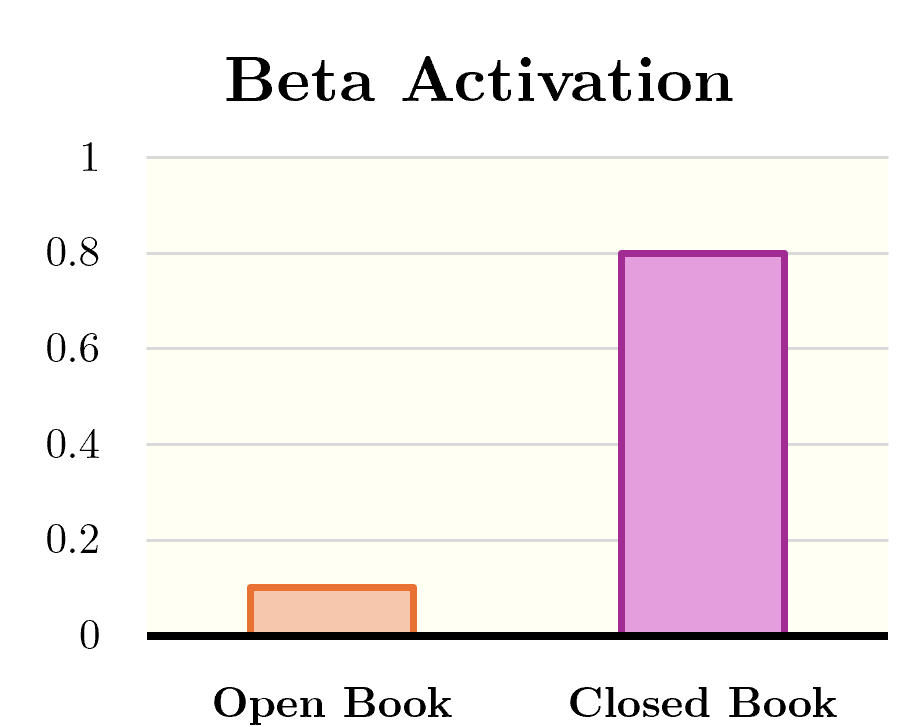}
    \caption{Beta Activation}
    \label{fig:pkm_activation}
  \end{subfigure}
  \caption{PKM factual recall ablation on synthetic QA probes. We compare open-book queries (fact present in prompt) with closed-book queries (fact omitted but stored in PKM). Metrics: (a) accuracy (fraction correct), (b) inference latency (ms/token), and (c) average PKM gate activation $\beta$. Trend: Hydra selectively activates PKM for closed-book cases, boosting accuracy while maintaining low latency, and suppresses PKM in open-book queries.}
  \label{fig:pkm_comparison}
\end{figure}

We created the dataset by generating synthetic QA pairs where a supporting fact is either included directly in the prompt (\emph{open-book}) or omitted but stored in PKM (\emph{closed-book}) \citep{sun2023recitation}. As shown in Figure~\ref{fig:pkm_comparison}, Hydra activates PKM ($\beta > 0$) primarily in the closed-book setting with a beta activation of $0.8$, while suppressing activation when the fact is already present with a $0.1$ beta activation for the open-book setting.

The above experiment demonstrates that Hydra’s Product-Key Memory (PKM) is not always active, but instead triggered selectively when facts are missing from the prompt. This matters because it shows that factual recall can be modularized rather than entangled with parametric weights. In practice, such selectivity allows Hydra to reduce unnecessary memory lookups, reducing latency while still recovering factual knowledge when needed. This property is critical for scaling reasoning systems to large knowledge bases without incurring constant retrieval overhead.

\subsection{Sparse Attention: Distant Premises}
\label{sec_SA:DP}
\begin{figure}[h]
  \centering
  \begin{subfigure}{0.32\linewidth}
    \includegraphics[width=\linewidth]{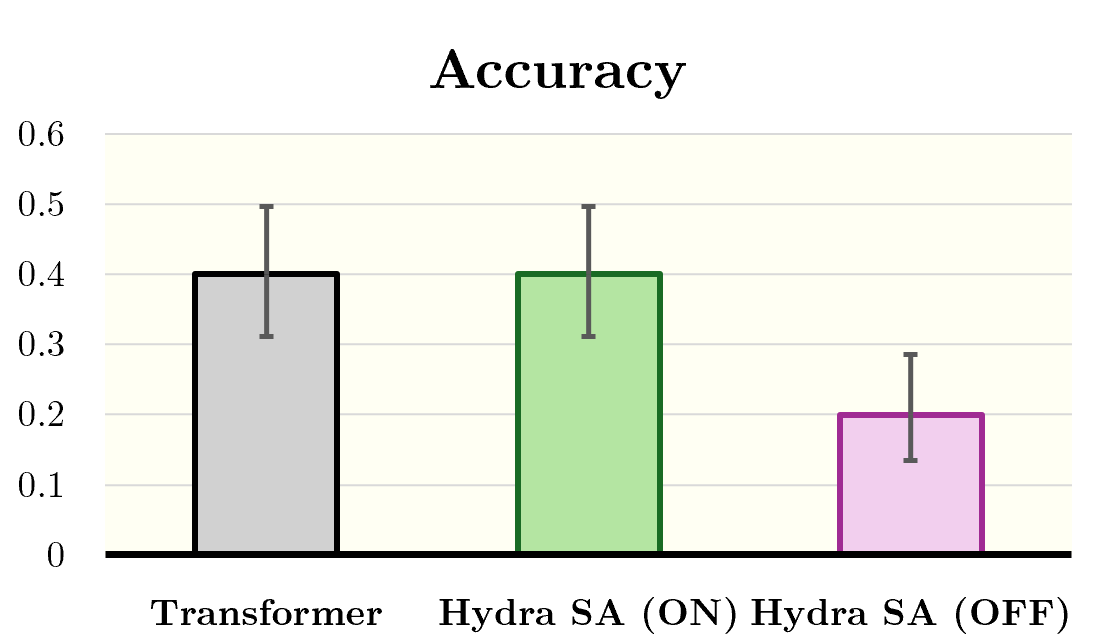}
    \caption{Accuracy}
    \label{fig:sparse_tokens}
  \end{subfigure}\hfill
  \begin{subfigure}{0.32\linewidth}
    \includegraphics[width=\linewidth]{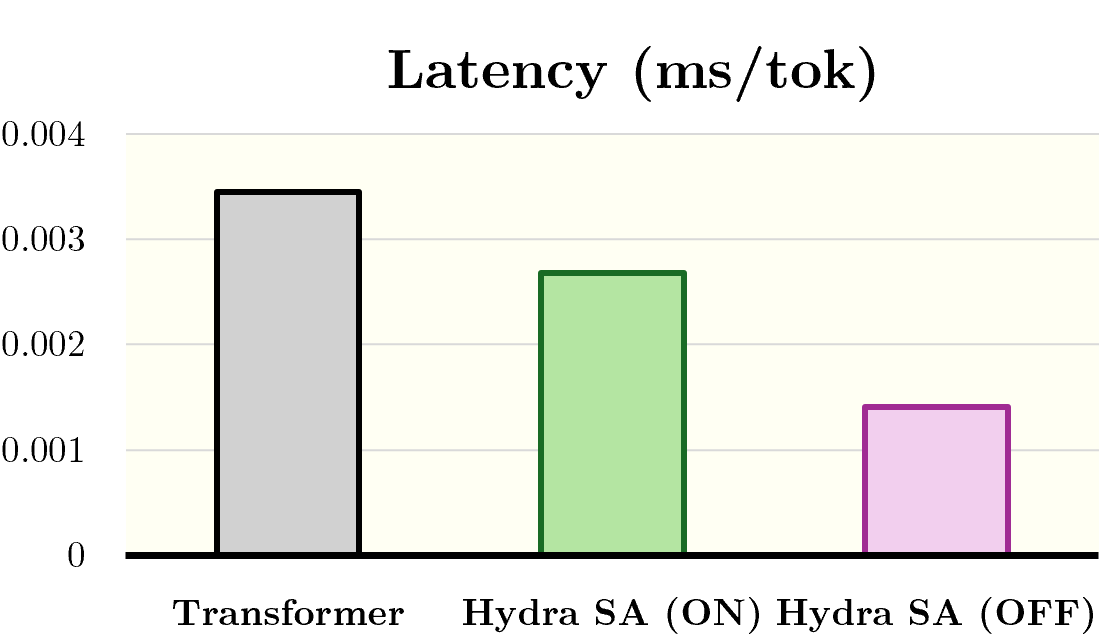}
    \caption{Latency}
    \label{fig:sparse_latency}
  \end{subfigure}\hfill
  \begin{subfigure}{0.32\linewidth}
    \includegraphics[width=\linewidth]{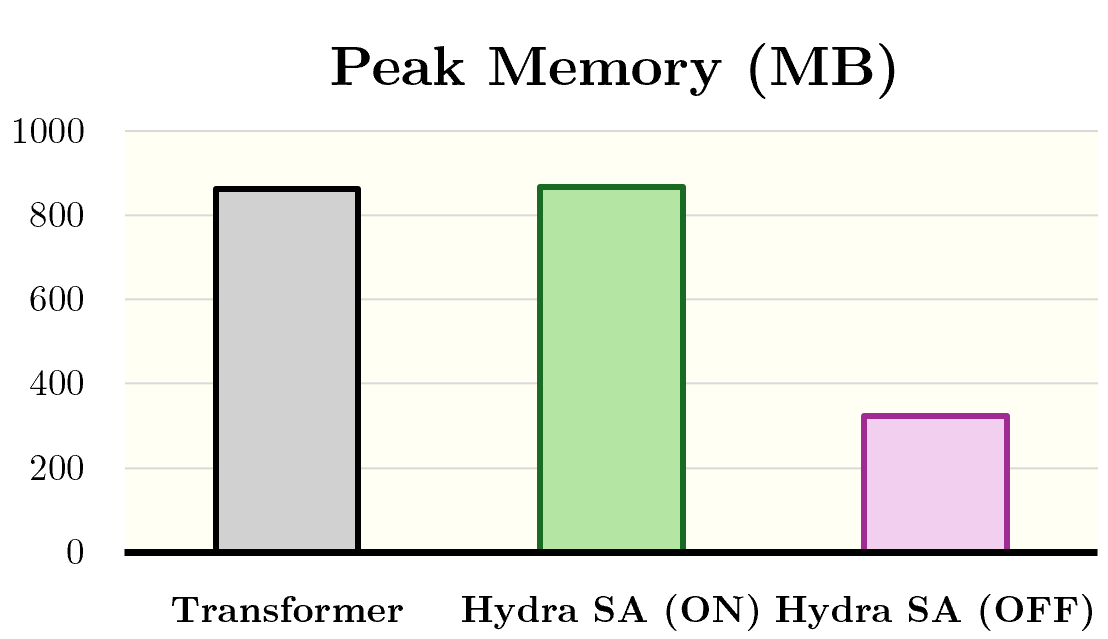}
    \caption{Tokens per second}
    \label{fig:sparse_accuracy}
  \end{subfigure}
  \caption{Sparse attention ablation on synthetic premise–conclusion tasks with long distractors. Metrics: (a) accuracy (conclusion prediction rate), (b) inference latency (ms/token), and (c) peak GPU memory (MB). Trend: Removing sparse attention severely reduces accuracy when premises are distant. Hydra with sparse global attention restores accuracy at far lower latency and memory cost compared to dense attention.}
  \label{fig:sparse_comparison}
\end{figure}

We created the dataset by generating sequences where a premise and its conclusion are separated by long distractor tokens, forcing the model to recover dependencies across distant spans. Without sparse attention, accuracy collapses. Figure~\ref{fig:sparse_tokens} shows that sparse attention maintains accuracy at 0.4 on par with the transformer baseline, but with 20\% less latency as shown in Figure ~\ref{fig:sparse_latency}.

Sequences of length $4096$ contained a single premise at $\sim2000$ and a query at the end; models had to recover the correct token (e.g., the premise’s color) across thousands of distractors, directly testing long-range dependency. Training used AdamW ($\text{lr}=10^{-3}$, batch size $1$) for 100 epochs, with five runs for stability, comparing a baseline transformer, Hydra without sparse attention, and Hydra with sparse attention under identical conditions.

These findings validate sparse attention as a lightweight mechanism for restoring
long-range dependencies that SSMs alone cannot capture. By achieving accuracy comparable to dense attention with less latency, Hydra ensures that key information can be recovered across long contexts without sacrificing efficiency or time, a property essential for reasoning over long documents or multi-hop inputs.

\subsection{MoE vs.\ Dense: Conditional Compute}
\label{sec_MoEvsDense}

\begin{figure}[h]
  \centering
  \begin{subfigure}{0.32\linewidth}
    \includegraphics[width=\linewidth]{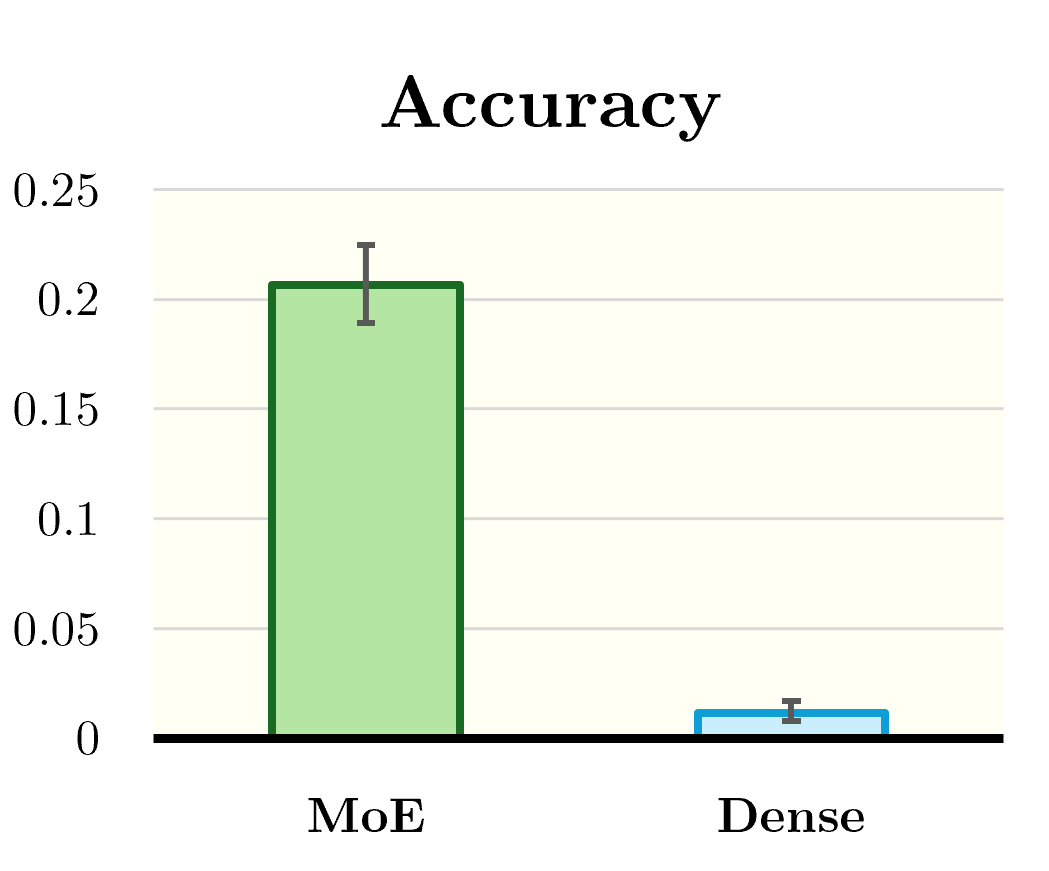}
    \caption{Accuracy}
    \label{fig:moe_tokens}
  \end{subfigure}\hfill
  \begin{subfigure}{0.32\linewidth}
    \includegraphics[width=\linewidth]{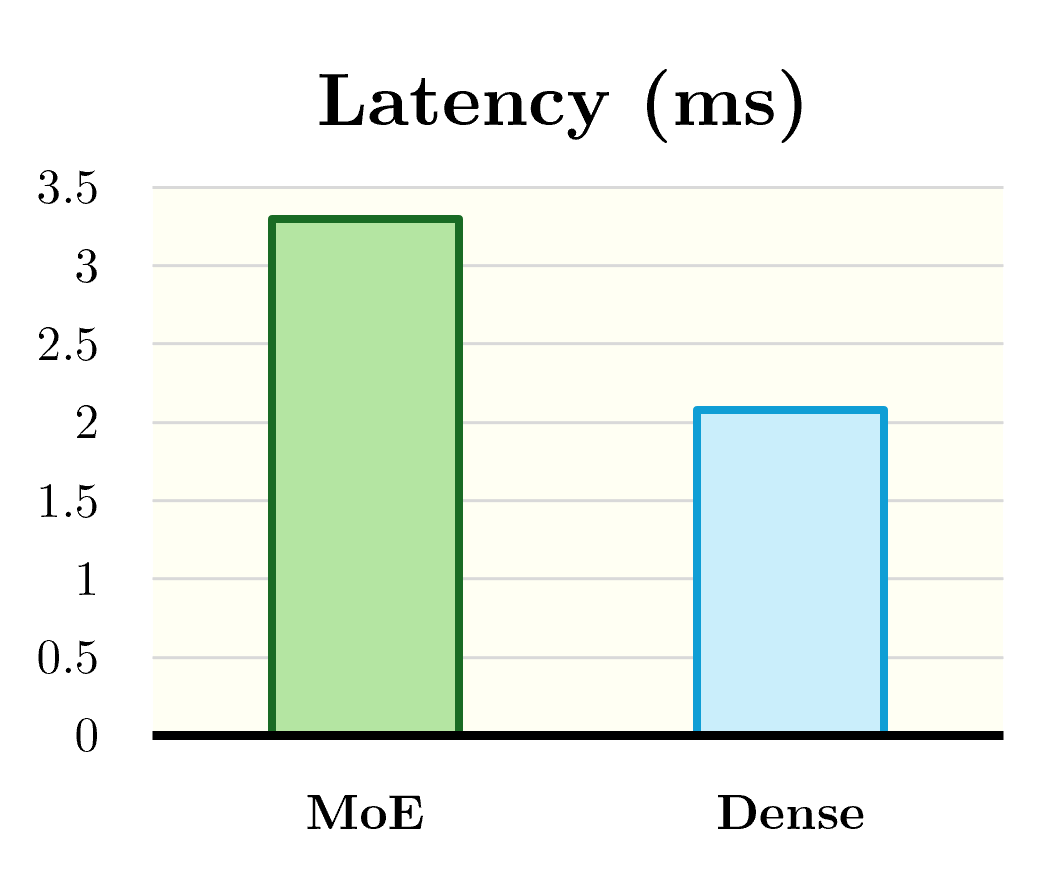}
    \caption{Latency}
    \label{fig:moe_latency}
  \end{subfigure}\hfill
  \begin{subfigure}{0.32\linewidth}
    \includegraphics[width=\linewidth]{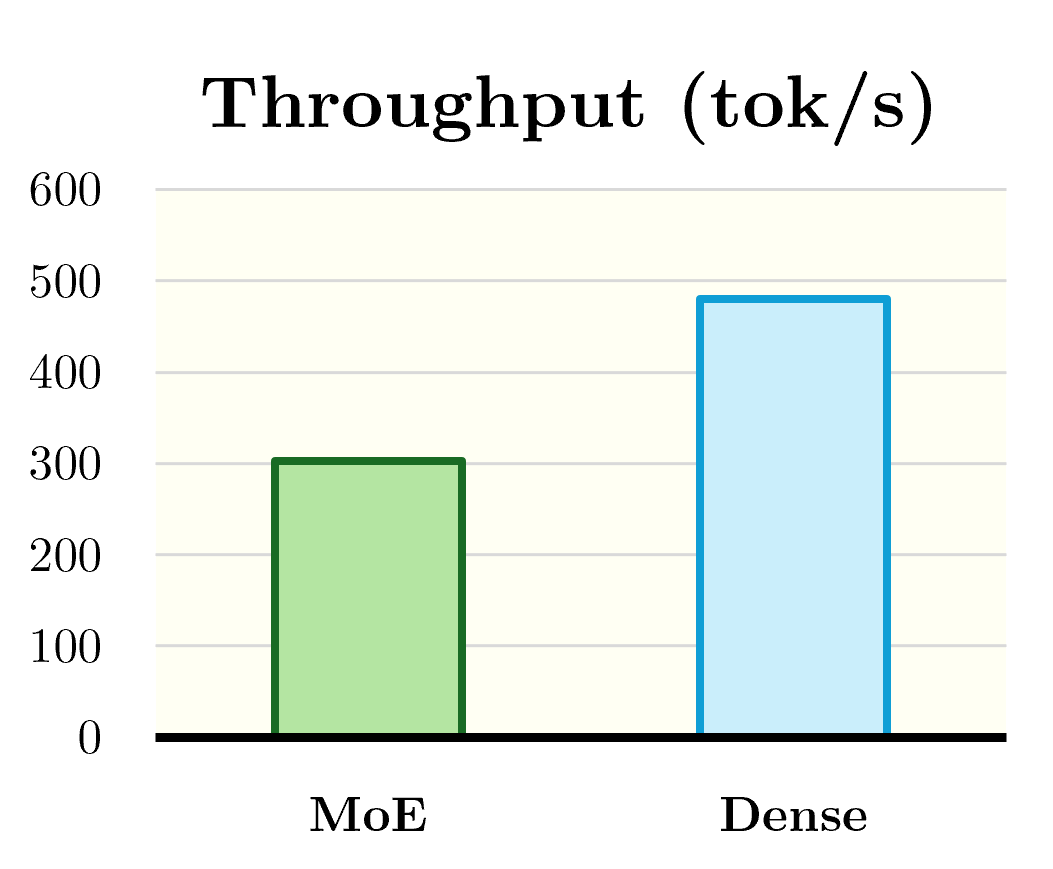}
    \caption{Tokens per second}
    \label{fig:moe_accuracy}
  \end{subfigure}
  \caption{MoE vs dense feed-forward comparison on synthetic multi-domain language modeling. Metrics: (a) accuracy (token prediction), (b) inference latency (ms/token), and (c) throughput (tokens/sec). Trend: Hydra’s chunk-level Top-2 MoE yields large accuracy gains from expert specialization, with only modest increases in latency and throughput trade-offs relative to a dense baseline.}
  \label{fig:moe_comparison}
\end{figure}

We compare Hydra with chunk-level MoE to dense FFN on synthetic multi-domain language modeling. 
The dataset is constructed by synthesizing several token distributions, each corresponding to a distinct \emph{domain} (for example, different alphabets or numeric ranges). 
Training sequences are sampled from a mixture of these domains, requiring the model to recognize domain identity and apply domain-specific processing. 
Figure~\ref{fig:moe_comparison} shows that the MoE variant improves accuracy by a factor of 10 through the specialization of experts at modest latency cost. Expert specialization observed here aligns with multi-domain benchmarks, again supporting synthetic tasks as effective stand-ins.

These results are useful because they demonstrate that chunk-level MoE routing enables domain-specific
expert specialization while keeping compute overhead modest. This conditional scaling of capacity means Hydra
can adapt to diverse input distributions (e.g., language, math, code) without inflating inference cost,
highlighting the practicality of modular compute allocation in real-world settings.

Models were trained for 5 epochs on 2000 synthetic arithmetic samples using AdamW (lr = 0.001). Evaluation covered the full dataset with randomized input order; accuracy was the fraction of correct answers, and inference latency/throughput were averaged per query across all samples.

\paragraph{Summary}  
Across ablations, Hydra’s modules behave as intended: workspace supports multi-step reasoning, PKM activates selectively, sparse attention enables efficient long-range lookups, and MoE expands conditional capacity. These signals confirm the feasibility of the architecture at toy scale.

\section{Conclusion}

Hydra introduces a modular architecture for efficient long-context reasoning by combining an SSM backbone with sparse attention, mixture-of-experts, and dual memories under a lightweight router. Our toy-scale results show that each component contributes distinct capabilities: workspace memory supports multi-step inference, PKM enables selective factual recall, sparse attention restores long-range dependencies, and MoE expands conditional capacity. Together, these yield both higher reasoning accuracy for logical chains by up to an order of magnitude and over $3\times$ throughput improvements compared to transformers of the same size at long sequence lengths. Rather than a single mechanism, our findings suggest that modularity is largely beneficial for balancing efficiency with reasoning depth.

\section{Discussions}

Our experiments are conducted at a toy scale to isolate component behaviors and validate feasibility. While this setup highlights how each module contributes in controlled conditions, it remains to be seen how these dynamics translate to billion-parameter models where routing stability, expert balance, and memory utilization may differ. The benefits of conditional computation, for instance, are expected to be even more pronounced at scale, but this has not yet been verified. Similarly, our memory modules behaved as intended in synthetic tasks, but their usage patterns in large, real-world deployments are still to be explored. Finally, Hydra has not yet been tested on competitive reasoning benchmarks, making such evaluations a natural next step to assess end-task effectiveness.

Scaling Hydra into the billion-parameter regime is a natural next step, enabling us to test whether the efficiency–reasoning trade-offs observed at toy scale hold in realistic settings. Such scaling would allow evaluations on diverse benchmarks spanning mathematics, code, and multi-hop QA, providing clearer evidence of Hydra’s end-task capabilities. Beyond scaling, several directions appear especially promising: (i) optimizing retrieval and memory kernels for real hardware to further reduce latency; (ii) experimenting with finer-grained or adaptive expert routing to strengthen conditional compute; (iii) integrating Hydra’s router with retrieval-augmented pipelines to couple modular compute with external knowledge; and (iv) extending the blueprint beyond text to multimodal reasoning in domains such as vision and speech. By keeping the architecture modular and transparent, we envision Hydra as a platform for the community to systematically explore design trade-offs and advance efficient reasoning.

\newpage

\bibliographystyle{plainnat}
\bibliography{references}


\appendix

\section{Appendix}

This appendix provides detailed specifications, parameter accounting, and training curriculum design for Hydra. These details are included for reproducibility and transparency, but are not required to follow the main results.

\subsection{Reference Configuration}
We illustrate Hydra with a $\sim$1.6B parameter instantiation. Numbers are illustrative and tunable via model dimension, MoE count, or PKM configuration.

\begin{itemize}
    \item \textbf{Tokenizer:} 50k BPE with tied input/output embeddings.
    \item \textbf{Model dimension:} $d=2048$.
    \item \textbf{Blocks:} 24 blocks arranged into 8 tri-path triples.
    \item \textbf{Context length:} native 16k; extended to 64k via workspace compression.
    \item \textbf{Attention (SGA):} every third block; local window $w=256$; up to $K=512$ router-selected global tokens. Typical on-rate $p_{\text{SGA}} \in [0.15,0.45]$.
    \item \textbf{MoE:} 12 alternating blocks with 6 SwiGLU experts each; Top-2 routing at 64-token chunk granularity.
    \item \textbf{Workspace memory:} 256 slots, with $\leq 64$ active per segment. Integration via low-rank factorized cross-attention.
    \item \textbf{Product-Key Memory (PKM):} $256 \times 256$ composite keys, value dim 1024, Top-4 retrieval.
    \item \textbf{Router + Retriever:} $\sim$40M parameters, operating on chunk-level summaries for conditional gating and retrieval.
\end{itemize}

\subsection{Parameter Accounting}\label{app:param-accounting}
Approximate trainable parameter counts are shown below.

\begin{center}
\begin{tabular}{l r}
\toprule
Component & Parameters \\
\midrule
Embeddings (tied) & 102.4M \\
24 SSM blocks (22.0M each) & 528.0M \\
8 Sparse Global Attention layers & 134.2M \\
12 MoE expert pools (6 $\times$ 9.44M each) & 679.5M \\
Workspace memory + mixers & 30.0M \\
PKM (keys, values, projections) & 69.8M \\
Retriever + Router & 40.0M \\
Output layers, norms, glue & 25.0M \\
\midrule
\textbf{Total} & \textbf{1609M} \\
\bottomrule
\end{tabular}
\end{center}

\paragraph{Active parameters.}  
In a typical forward pass, $\sim$0.80--0.83B parameters are active: 528M (SSM) + 226.6M (MoE, Top-2 experts) + $p_{\text{SGA}} \cdot 134.2$M (attention) + $\sim$25M glue. Sparse pathways (attention + experts + memory) expose $\sim$250--305M conditional weights depending on $p_{\text{SGA}}$.

\subsection{Product-Key Memory Lookup}
PKM uses two sub-key codebooks $K^{(1)}, K^{(2)} \in \mathbb{R}^{256 \times 128}$ to form $65{,}536$ composite keys. A query $q_t \in \mathbb{R}^{256}$ is split into $(q^{(1)}, q^{(2)})$, each side retrieves $t=8$ nearest sub-keys. Candidate composite keys are formed from the Cartesian product ($t^2=64$), scored as
\[
s_{ij} = \langle q^{(1)}, K^{(1)}_i \rangle + \langle q^{(2)}, K^{(2)}_j \rangle.
\]
Top-$K_c=4$ composites are selected, values aggregated as
\[
m_t = \sum_{(i,j)} \alpha_{ij} V_{ij}, \quad \alpha_{ij} = \mathrm{softmax}(s_{ij}).
\]
A learned gate $\beta_t$ blends retrieved values into the hidden state: 
\[
h_t \leftarrow h_t + \beta_t W_{val} m_t.
\]

\subsection{Workspace Memory}
Workspace memory consists of 256 learnable slots $M = \{m_i\}$. At most 64 slots are active per segment. Tokens can \emph{write} chunk summaries into slots and \emph{read} from updated slots via low-rank projections ($r=256$). For extended contexts, active slots are compressed into 64-slot summaries passed across segments.

\subsection{Complexity Summary}
For sequence length $T$ and dimension $d$:
\begin{itemize}
    \item SSM path: $O(Td)$.
    \item Sparse attention: $O(T(w+|G|)d)$.
    \item MoE: Top-2 experts per chunk $\Rightarrow$ FLOPs/token $\sim O(dh)$ with $h < d$.
    \item Workspace: $O((T+S)r)$ with $S \leq 64$.
    \item PKM: $O(d d_k + t^2)$, dominated by key projection and candidate scoring.
\end{itemize}

\subsection{Training Curriculum (Phases A--D)}
Hydra’s conditional components are activated in stages to stabilize optimization:
\begin{enumerate}
    \item \textbf{Phase A:} Train backbone (SSM + embeddings) only.
    \item \textbf{Phase B:} Introduce sparse attention at low on-rate; gradually increase activation.
    \item \textbf{Phase C:} Enable MoE layers with auxiliary load-balancing loss; anneal expert dropout.
    \item \textbf{Phase D:} Activate workspace and PKM memories with gating losses and retrieval supervision.
\end{enumerate}
This curriculum is forward-looking: toy-scale prototypes were trained without full curriculum activation, but staged scheduling is expected to be necessary at scale.

\end{document}